%% file: main.tex
\documentclass[10pt]{article}

\usepackage[T1]{fontenc}
\usepackage[utf8]{inputenc}
\usepackage{lmodern}

\usepackage{graphicx} 
\usepackage{xcolor}
\usepackage{booktabs}
\usepackage{colortbl}
\usepackage{hyperref}
\usepackage{multirow}
\usepackage[noend]{algorithm2e}
\SetAlFnt{\small}
\SetAlCapFnt{\scriptsize}

\usepackage{adjustbox}
\usepackage{tablefootnote}

\usepackage{amsmath,amssymb,amsfonts}
\usepackage{amsthm}
\usepackage{mathrsfs}
\usepackage[title]{appendix}
\usepackage{textcomp}
\usepackage{tikz}
\usetikzlibrary{arrows.meta,fit,backgrounds,calc}

\usepackage[square,sort,comma,numbers]{natbib}

\title{Two-dimensional early exit optimisation of LLM inference}

\author{
Jan Hůla$^{1}$ \  David Adamczyk$^{1}$ \  Tom\'a\v s Filip$^{1}$ \  Martin Pavl\'\i\v cek$^{1}$\thanks{Corresponding author} \  Petr Sos{\'\i}k$^{1,2}$\\[0.5em]
$^{1}$Institute for Research and Applications of Fuzzy Modelling, \\ University of Ostrava, 
70200 Ostrava, Czech Republic\\[0.5em]
$^{2}$Institute of Computer Science, Faculty of Philosophy and Science,\\ Silesian University in Opava,
74601 Opava, Czech Republic\\[0.5em]
\normalsize\texttt{\{jan.hula, david.adamczyk, tomas.filip, martin.pavlicek, petr.sosik\}@osu.cz}
}
\date{}

\begin{document}
\maketitle

\begin{abstract}
We introduce a two-dimensional (2D) early exit strategy that coordinates layer-wise and sentence-wise exiting for classification tasks in large language models. By processing input incrementally sentence-by-sentence while progressively activating deeper layers, our method achieves multiplicative computational savings that exceed those from optimizing either dimension independently. Experimental evaluation across four state-of-the-art LLMs (Llama 3.1, Llama 3.2, Gemma, Qwen; 3B-8B parameters) on three sentiment classification datasets demonstrates additional speed-ups of 1.4--2.3$\times$ over optimal layer-wise early exit for simpler tasks with vanilla models, with graceful degradation on complex multi-class problems. Fine-tuning reduces but does not eliminate this advantage. The approach is model-agnostic, requires only lightweight classification adapters, and is orthogonal to complementary efficiency methods such as quantization and pruning. Our findings indicate that 2D early exit strategies excel when semantic information accumulates predictably across input structure, suggesting possible applicability to sequence-processing tasks beyond sentiment classification.

\end{abstract}


\section{Introduction}
\label{sec_intro}
Large Language Models (LLMs) have revolutionised Natural Language Processing (NLP) with their remarkable capabilities across a range of tasks. However, their extraordinary performance comes at a significant computational cost in terms of processing power, memory, and energy. This applies especially to the inference regime, which is run extensively once the model is trained, while training is a very demanding but usually one-time task. For relatively simpler but high-throughput tasks such as classification, 
deploying full-scale LLMs represents an immense computational overhead. 
To this end, early exit strategies have emerged as a promising paradigm. They enable a model to terminate its computation at an intermediate layer when it is sufficiently confident in its prediction, while striving to maintain high accuracy. 

\begin{figure}
    \centering
    Gemma-3n-E4B vanilla (adapter-only training)

    \includegraphics[width=\linewidth,trim=0 0 0 2.0cm,clip]{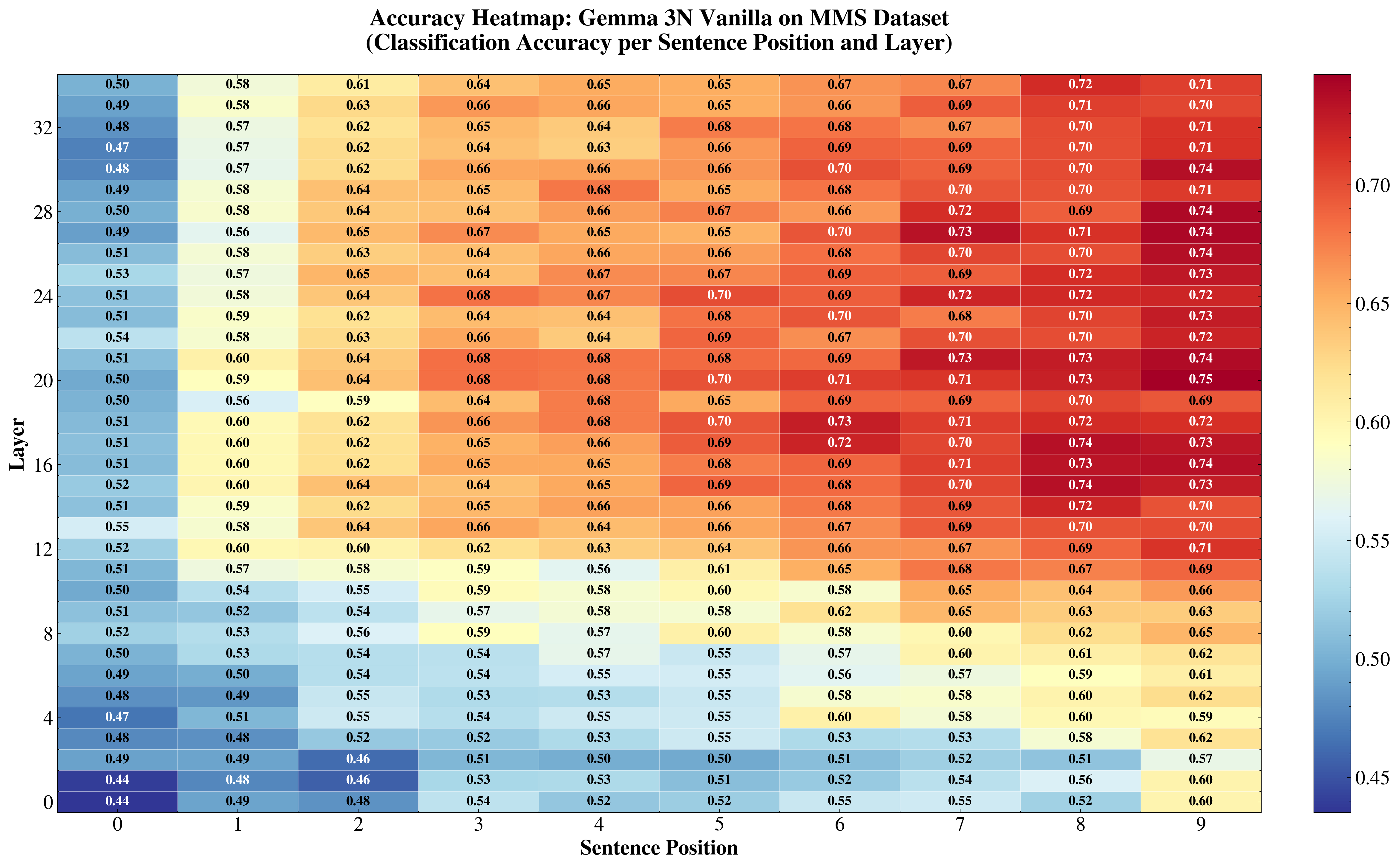}

    Gemma-3n-E4B fine-tuned

    \includegraphics[width=\linewidth,trim=0 0 0 2.0cm,clip]{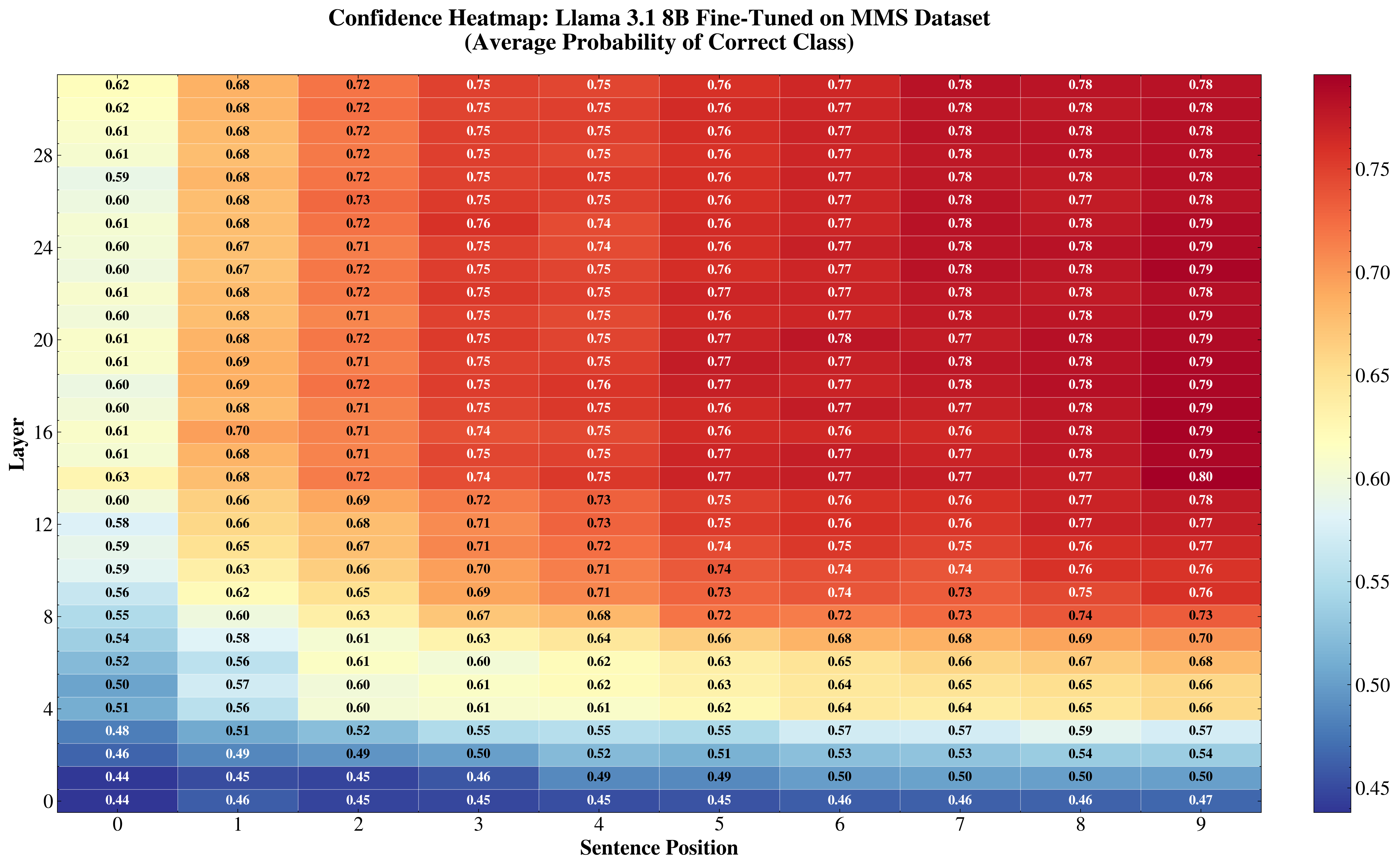}
    \caption{Layer-wise and sentence-wise accuracy of correct classification of Gemma-3n-E4B vanilla (top) and fine-tuned (bottom) on the MMS dataset. Heatmaps based on 7100 samples of  10 sentences each.}
    \label{fig:gemma_heatmap_sentences}
\end{figure}

In this paper, we introduce a two-dimensional (2D) early exit optimisation strategy that combines layer-wise and sentence-wise early exit. The intuition behind it is that, in many classification tasks, the proper class can already be determined from an initial part of the text. To preview the anatomy of the 2D EE strategy, Figure \ref{fig:gemma_heatmap_sentences} demonstrates how the probability of correct classification progresses along sentences and layers during document processing.
Taking a sentence as a natural semantic unit, we split an input sample into sentences and feed an LLM input with sentences one-by-one. Simultaneously, with each new sentence we increment the number of layers processing the so-far read part of the input.
This results in rectangular processing blocks growing in 2D (layers vs. sentences) illustrated in Fig. \ref{fig:chunks}. The classification halts once the accumulated confidence (\ref{eq_confidence}) exceeds a pre-defined threshold. In this way, a significant speed-up can be achieved compared to standard layer-wise early exit methods. Our findings indicate a possible applicability to a range of sequence-processing tasks, where each input sample can be meaningfully split into a sequence of semantic units that contribute to the overall meaning in an incremental manner. The main contributions are:
\begin{itemize}
\item \textbf{Methodology:} We integrate layer-wise early stopping with sentence-wise input trimming into one synergistic process. By simultaneously coordinating exiting in both dimensions we achieve multiplicative rather than additive computational savings without sacrificing accuracy.

\item \textbf{Benchmarking:} We evaluate the 2D strategy against optimal layer-wise early exit and LayerSkip across four state-of-the-art LLMs (Llama-3.1-8B, Llama-3.2-3B, Gemma-3n-E4B, Qwen2.5-7B) on three sentiment classification datasets of varying difficulty. The results demonstrate 1.4--2.3$\times$ additional speedup (relative to layer-based early exit) for simpler tasks in vanilla models, with graceful degradation on harder problems.

\item \textbf{Analysis and insights:} We analyse how fine-tuning affects the 2D advantage, provide hyperparameter optimisation guidance, and identify overall conditions for 2D strategies to excel.
\end{itemize}

The rest of the paper is structured as follows. Section~\ref{sec_background} reviews related work. Section~\ref{sec_methods} outlines the methodology used, and Section~\ref{sec_experiments} resumes and analyses the experimental results. Section~\ref{sec_discussion} discusses the implications of the work, and Section~\ref{sec_conclusion} concludes the paper with directions for future extensions.

\section{Background}
\label{sec_background}

Traditional methods to improve inference efficiency in LLMs include pruning \cite{ma2023llm}, weight quantisation \cite{dettmers2023qlora}, and knowledge distillation \cite{jiao2020tinybert}. These techniques compress the model statically; that is, the resulting model operates with fixed structure regardless of input. Although effective, they often cause accuracy degradation and may overfit to specific datasets or domains, reducing transferability between tasks. 
In contrast, dynamic inference methods adapt computation at runtime.  
A classical dynamic approach is to employ a \textit{cascade} of models with increasing complexity, where a lightweight model handles easy instances, while only harder ones are passed to a larger model \cite{chen2023frugalgpt}. The \textit{Matryoshka} paradigm \cite{NEURIPS2022_matryoshka} generalises this idea into a cascade of progressively complex nested models. MatFormer \cite{devvrit2024matformer} integrates feedforward network blocks within transformer layers, allowing the extraction of multiple smaller models from one training process at no additional cost.  

\subsection{Early exit strategy}

Early exit methods reduce inference cost by terminating the computation in intermediate layers once a predefined criterion (the \textit{exit trigger}) is met. They are complementary to pruning, quantisation, or distillation and are applicable to virtually any transformer model without fundamental architectural change. The foundational paper \cite{teerapittayanon2016branchynet} introduced the early exit paradigm for deep neural networks. For an overview of the topic, we refer, e.g., to the survey \cite{laskaridis2021adaptive} or \cite{rahmath2024early}.

\medskip\noindent
\textbf{Confidence-based exits.} Early works attach auxiliary classifiers to intermediate layers and allow inference to stop once confidence exceeds a threshold. DeeBERT \cite{xin2020deebert} pioneered entropy-based exits (off ramps) for BERT, achieving up to $40\%$ inference time reduction on GLUE dataset benchmarks with a drop in accuracy $<4\%$. Schwartz et al. \cite{schwartz2020right} proposed calibrated confidence thresholds, showing that multi-exit BERTs could achieve $3$--$5\times$ faster inference with almost the same accuracy. To strengthen shallow exits, FastBERT \cite{liu2020fastbert} applied self-distillation and adaptable inference, producing reliable intermediate predictions and allowing tunable $1$--$12\times$ speed-ups with graceful accuracy degradation. PABEE \cite{zhou2020bert} introduces the exit strategy based on an unchanged prediction of internal classifiers for a predefined number of consecutive steps.

\medskip\noindent
\textbf{Learned exit policies.} Rather than fixed heuristics, some approaches learn when to exit. Li et al. \cite{li2023predictive} introduced a predictive engine for early exiting decisions that allows the network to forecast when it should exit, adapting to computational and energy constraints in real-time. ConsistEE \cite{zeng2024consistenteeconsistenthardnessguidedearly} employs reinforcement learning to align training-time supervision with inference-time exits, reporting double digit percentage layer depth reductions with the same or better accuracy. Schuster et al. \cite{schuster2021consistent} proposed Confident Adaptive Transformers (CAT), combining additional prediction heads with meta consistency classifiers to guarantee confidence, while still obtaining $1.8$--$2.7\times$ acceleration. Fan et al. \cite{fan2024not} uses statistical classifiers such as SVM trained on easily obtainable token-based features to adaptively decide when to stop inference.

\medskip\noindent
\textbf{Architecture-integrated approaches.} Some methods modify training or architecture to support the exit procedure. LayerSkip \cite{Elhoushi_2024} introduces layer dropout with progressively higher dropout rates for deeper layers. The early exit loss supervises all intermediate layers using a weighted combination. Self-speculative decoding enables flexible skipping, resulting in $1.82$--$2.16\times$ speedup on coding, parsing and summarization tasks with marginal accuracy loss. Liu et al. \cite{liu2024kangaroo} uses a fixed shallow sub-network as a self-draft model with a controlled exit, while the remaining layers serve as the larger target model. GREEN-CODE \cite{ilager2025greencodeoptimizingenergyefficiency} uses reinforcement learning to optimise exit policy in code generation, demonstrating up to $23$--$50\%$ energy savings in Llama-3.2-3B and OPT 2.7B. Chen et al. \cite{chen2023ee} introduces a variety of optimisations targeted both in the training and inference phase, enabling early-exit LLMs to scale up to 30B parameters.

\medskip\noindent
\textbf{Token-level and sequence-level extensions.} Many works, including some already cited \cite{liu2020fastbert,Elhoushi_2024}, combine layer-wise early exit with token-wise exiting, where different tokens exit at different layers. At the sequence level, token-wise halting was pioneered by the Depth-Adaptive Transformer \cite{elbayad2020depth}, which achieved faster decoding in machine translation. CALM (Confident Adaptive Language Modeling) \cite{schuster2022calm} introduced dynamic allocation of different amounts of compute per input and achieved up to $3\times$ speed-up on generative tasks. Khanna et al. \cite{ee-silver} presented a model-agnostic inference framework that reduces computation via dynamic token halting, combined with KV-cache skipping and token-level redundancy reduction, achieving up to 40\% resource reduction on frozen models. Recent work extends early exit to reasoning tasks: DEER \cite{ee-deer} monitors model confidence at CoT transition points to terminate reasoning sequences early (19--80\% reduction in CoT length), while NEAT \cite{liu2026neat} monitors neuron-level activation dynamics for fine-grained exit decisions (22--28\% token reduction across CoT reasoning benchmarks).

Our 2D early exit strategy shares with CALM \cite{schuster2022calm} and token-halting approaches \cite{elbayad2020depth,ee-silver} the intuition that not every input unit warrants equal computational depth. However, these methods allocate variable compute per token within a fixed context, requiring careful handling of absent hidden representations and KV-cache management. By contrast, operating at the sentence level -- a natural semantic boundary -- ensures well-defined embeddings at every active layer, avoiding these complications and leveraging the predictable accumulation of sentiment signal across sentences. Furthermore, by coordinating sentence-wise input trimming with layer-wise early stopping within a unified framework, we achieve multiplicative rather than additive resource reductions in classification tasks.


\section{Methods}
\label{sec_methods}

The core idea of 2D early exit in transformers is to combine layer-wise and token-wise exiting into one framework that acts simultaneously in both dimensions to obtain a synergistic effect. We have chosen a sentence as the elementary input semantic unit. Each sample in the dataset is split into sentences by the algorithm detailed in the Appendix \ref{app_sentence_splitting}. Throughout the paper, \textit{sentence embedding} is defined as the mean of the embeddings of all tokens in the sentence. 
Each tested model is supplemented with lightweight classification adapters (or classifiers, for short), one per layer, fed with the output embedding of that layer. The classifier architecture is a two-layer fully connected network:
\begin{itemize}
\item \texttt{Linear(embedding\_dim $\rightarrow$ 256) + RelU}
\item \texttt{Linear(256 $\rightarrow$ num\_classes) + Softmax}
\end{itemize}

\subsection{Training}
\label{sec:training}

The loss function used for training is based on that used in \cite{xin2020deebert, zhou2020bert, Elhoushi_2024} that aims at accurate predictions at intermediate layers. Simultaneously, the loss function was expanded to promote correct classification based on a partial input. Consider a model with $L$ transformer layers. For a sample $(x,y)$ consisting of $m$ sentences, denote by $e_{i,k}(x)$ the embedding of sentence $k,$ $0\le k\le m-1,$ of sample $x$ in the output of layer $i,$ $0\le i\le L-1.$

Let us define the \textit{prefix embedding} $pe_{i,j}(x)$ as the mean embedding of the first $j$ sentences of the sample,

\begin{equation}\label{eq_prefix_embedding}
    pe_{i,j}(x) =  \sum_{k=0}^{j-1} \dfrac{e_{i,k}(x)}{j},\quad 0\le j\le m-1.
\end{equation}
Two variants of training have been tested: (i)  adapter-only training and (ii) fine-tuning of the whole model. 

\subsubsection*{Adapter-only training}

In this variant, only the classifiers for individual layers are trained, while the model weights remain frozen. For a training sample $(x,y)$  of $m$ sentences, each classifier is optimised to predict the target label $y$ solely based on the output embedding of the corresponding layer. The sentence embeddings are aggregated using the mean operation and finally processed by the classification adapter, resulting in the loss function
\begin{equation}\label{eq_layer_loss}
    \mathcal{L}_i(x,y) = \mathrm{CE}\Big(y,\mathrm{fc}_i\Big(\sum_{j=0}^{m-1}  \dfrac{e_{i,j}(x)}{m}\Big)\Big),
\end{equation}
where $\mathrm{fc}_i(\cdot)$ is the output of the $i$-th classification adapter (class probability distribution) and $\mathrm{CE}(\cdot,\cdot)$ is the cross-entropy function. 

\subsubsection*{Fine-tuning}
    
The fine tuning of all tested models, including the classifiers, is implemented with the PEFT adapter. The aggregate loss function 
\begin{equation}\label{eq_FT_loss}
    \mathcal{L}(x,y) = \sum_{i=0}^{L-1} \lambda_{i}\ \mathrm{CE}\Big(y,\mathrm{fc}_i\Big(\sum_{j=0}^{m-1}  \dfrac{pe_{i,j}(x)}{m}\Big)\Big)
\end{equation}
for a sample $(x,y)$ is a weighted average of layer losses. Each layer loss is based on the output of the layer classifier, where its input is an average of all prefix embeddings (\ref{eq_prefix_embedding}) for this layer and sample. The weights $\lambda_{i}$ were experimentally set to 0.8 for the last layer $L-1$ and 0.1 for all previous layers. These values provided the best accuracy across the three tested datasets; however, they may be further auto-tuned for different datasets.
%
%

\subsection{Inference}
\label{subsec_inference}


\begin{figure}[t!]
    \centering
    \input{Schema_2D_EE}
    \caption{Visualization of the 2D early exit strategy for step size $\Delta=2,$ see eq. (\ref{eq_chunk_size}). Arrows  Red frames enclose inference progression blocks corresponding to individual sentences.}
    \label{fig:chunks}
\end{figure}
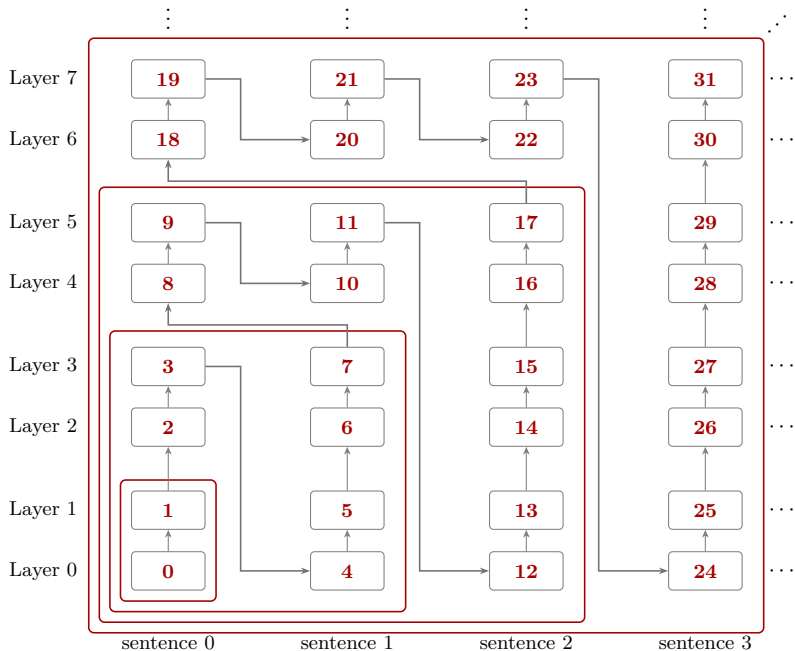

The key innovation of the 2D early exit strategy, detailed in Algorithm \ref{alg:early-exit}, is the two-dimensional progression: the algorithm processes each input sample sentence-by-sentence, with each new sentence progressively increasing the number of available network layers for inference. For an input sample with $m$ sentences, when processing a sentence indexed with $s$, $0\le s\le m-1,$ the algorithm activates the first $L_s$ layers of the model, 
\begin{equation}\label{eq_chunk_size}
L_s = \min((s+1) \times \Delta, L),\mbox{ where } \Delta = \max(1, \lfloor L / m \rfloor).
\end{equation}

All previous sentences $s_1 < s$ pass through newly activated layers they have not yet passed through, i.e., from $L_{s-1}$ to $L_s - 1,$ while sentence $s$ passes through all layers from 0 to $L_s - 1,$ resulting in red-framed blocks illustrated in Fig. \ref{fig:chunks}. Hence, the layer depth is coupled to the sentence index, so each new sentence simultaneously
expands both contextual breadth and representational dept 

Early stopping is controlled by two hyperparameters: ignore threshold $\tau_{ignore}$ and acceptance threshold $\tau_{acc}.$
Denote by $\mathrm{fc}_{\ell,c}(s_1)$ the softmax output (probability) of the classification adapter in layer $\ell$ for the input $s_1,$ corresponding to class $c.$
For each $\ell$ and $s_1,$ the algorithm computes \textit{confidence} as the difference between the maximum and second-maximum softmax outputs: 
\begin{equation}\label{eq_confidence}
\text{confidence} = \max\limits_{c}\{\mathrm{fc}_{\ell,c}(s_1)\} - \text{second\_}\max\limits_{c}\{\mathrm{fc}_{\ell,c}(s_1)\}. 
\end{equation}

\smallskip\noindent
If confidence exceeds $\tau_{ignore}$, it is added to the accumulated confidence score of the predicted class. When the accumulated confidence of any class exceeds $\tau_{acc},$ the inference terminates immediately, returning the prediction with the highest score. If no class reaches the $\tau_{acc}$  threshold after processing all layers and sentences, the algorithm returns the prediction of the final layer for the last sentence.

\begin{algorithm}
\SetAlgoLined
\LinesNumbered
\DontPrintSemicolon

\caption{The 2D early exit inference for one input example.}
\label{alg:early-exit}
\textbf{The 2D early exit inference pseudocode}
\\[0.5em]
\begin{tabbing}
\textbf{Input:}\qquad\= $example:$ input text \\
  \> $\tau_{ignore}, \tau_{acc}:$ hyperparameters -- ignore and acceptance threshold      \\
\textbf{Output:}\> $predicted\_label:$ prediction for the input example \\
  \> $operations\_used:$ consumed abstract operations to measure speed-up  
\end{tabbing}

\textbf{1. Initialization:}\\[0.5em]
  $sentences \gets$ list of sentences in \textit{example} \\
  $num\_sentences \gets$ number of sentences in \textit{example} \\
  $num\_layers \gets$ number of layers of the model \\
  $operations\_used \gets 0$ (counter of operations used) \\
  $\Delta \gets \max(1, num\_layers \text{ // } num\_sentences)$ (step size in layers) \\
  $acc \gets[0,0,0,...]$ (confidence accumulators for each class) 

\vspace{0.5em}
\textbf{2. Process sentences and layers till early stopping:}

\vspace{0.5em}
\For{$s = 0$ \textbf{to} $num\_sentences - 1$}{
  
  \vspace{0.5em}
  // Process all sentences up to $s$
    
  \For{$s_1 = 0$ \textbf{to} $s$}{
      
    \vspace{0.5em}
    // The number of active layers for sentences up to $s$
      
    $layers\_to\_traverse = \min((s+1) \times \Delta, num\_layers)$
      
    \vspace{0.5em}
    // Sentences before $s$ pass only newly added chunk of layers
      
    \textbf{if} $s_1 = s$ \textbf{then} $start\_layer = 0$ 
    
    \ \ \ \ \ \ \ \ \ \ \ \ \ \textbf{else} \ $start\_layer = \Delta * s$
      
    \vspace{0.5em}
    \For{$l = start\_layer$ \textbf{to} $layers\_to\_traverse - 1$}{

      $operations\_used \gets operations\_used + 1$

      \vspace{0.5em}
      // Calculate embeddings for the actual sentence using already stored 
      
      // embeddings for previous layers and sentences 
        
      $embeddings[l][s_1] \gets calculate\_embedding(l, sentences[s_1])$

      \vspace{0.5em}
      // Classifier at layer $l$ calculates probabilities for all classes
        
      $[p_0, p_1, p_2, \ldots] \gets \textit{classifier}_l(embeddings[l][s_1])$

      $predicted\_label \gets \arg\max([p_0, p_1, p_2, \ldots])$
      
      \vspace{0.5em}
      // Calculate confidence of the winning class (max - second max)
      
      $sorted\_probs \gets \text{SORT}([p_0, p_1, p_2, \ldots])$ descending
      
      $confidence \gets sorted\_probs[0] - sorted\_probs[1]$
            
      \vspace{0.5em}
      //  Accumulate confidence, filter by ignore threshold
      
      \If{$confidence > \tau_{ignore}$ }{ 
        $acc[predicted\_label] \gets acc[predicted\_label] + confidence$
        
        \vspace{0.5em}
        // Early stopping check
      
        \If{$acc[predicted\_label] > \tau_{acc}$}{
          \Return $(predicted\_label, operations\_used)$
        }
      }
    }
  }
}

\vspace{0.5em}
\textbf{3. All layers and sentences processed -- no early stopping:}

\vspace{0.5em}
$final\_label \gets$ prediction from the last sentence at last layer

\Return $(final\_label, num\_layers \times num\_sentences)$
\vspace*{2em}

\end{algorithm}

\subsection{Metrics and evaluation}
\label{sec_metrics}

The experiments were evaluated  using two metrics: (i) classification accuracy and (ii) speed-up enabled by early exit, expressed as the ratio of abstract operations in a full and early terminated inference. For each model, dataset, and training mode, let $acc(\ell)$ be the accuracy of the classifiers at layer $\ell$ on the test split. We set the acceptable \textit{accuracy threshold} to
\begin{equation}
\label{eq:threshold}
    \textit{acc\_thr} = \max\limits_{0\le \ell< L}\{acc(\ell)\} - T,
\end{equation}
where $T$ is the allowed accuracy loss. We repeated the experiments for $T=0.02$ (2\%) and $T=0.04$ (4\%). Then, for all EE methods tested, we ran the inference until the accuracy threshold was met and calculated the corresponding speed-up. For layer-wise early exit, the speed-up was set as 

\begin{equation}
\label{eq:reduction_layer}
\textit{speed-up} = L/(L_e + 1),   
\end{equation}
where $L$ is the total number of transformer layers in the model and $L_e$ is the exit layer at which the accuracy threshold is met (layers numbered from 0 to $L-1$).

The speed-up achieved by the 2D EE strategy was quantified as follows. We count the processing of one sentence in one transformer layer as one abstract operation. When the accuracy threshold is not met, Algorithm \ref{alg:early-exit} processes the complete input through all layers using $m \times L$ operations, which corresponds to an inference with no early exit. When early stopping occurs, the algorithm returns the number of \textit{operations\_used}, and the speed-up is calculated as
\begin{equation}
\label{eq:reduction_2D}
\textit{speed-up} = (m \times L) / \textit{operations\_used}.    
\end{equation}
One can argue that, since each sentence attends to all previous sentences, its processing time is proportional to its order in the input text. However, this is true only for long inputs of at least thousands of sentences. Denote

\medskip\noindent
\begin{tabbing}
    \textit{embed\_dim}\quad \= \kill
    \textit{tps} \>  average number of tokens per sentence ($\approx 15$ for user reviews);\\
    \textit{embed\_dim} \> embedding dimension of the model;\\
    \textit{exp\_f} \> expansion factor in the MLP module in the transformer layer. 
\end{tabbing}

\noindent
Considering, e.g., Llama-3.2-3B with $\textit{embed\_dim}= 3072$ and $\textit{exp\_f} \approx 2.67,$ computationally intensive operations in a transformer layer processing the $s$-th sentence of an input example, are:

\medskip\noindent
\begin{tabbing}
  MLP expansion and compression:\ \= \kill 
  Query, key, value computation:  \>  $\approx 3\times \textit{tps} \times \textit{embed\_dim}^2 \approx 4.2 \times 10^8 $\\
  Attention matrix computation:   \>  $\approx s\times \textit{tps}^2 \times \textit{embed\_dim} \approx 6.9\times 10^5\times s$\\
  MLP expansion and compression:  \>  $\approx 2\times \textit{tps} \times \textit{embed\_dim}^2 \times \textit{exp\_f} \approx 7.6 \times 10^8 $
\end{tabbing}

\noindent
Therefore, the linear term containing $s$ becomes relevant for $s = \Omega(10^3).$


\section{Experiments}
\label{sec_experiments}

The aim of the experiments was to compare the efficiency of the 2D EE strategy with recent approaches based on exiting at intermediate layers. Numerous studies, such as \cite{xin2020deebert,zhu2021leebert,schuster2021consistent,li2023predictive,Elhoushi_2024,fan2024not,liu2024kangaroo}, and many others, have examined methods of triggering layer-wise early exit. They often differ in language models and datasets, and some use additional mechanisms to improve early exit efficiency, which makes comparison difficult or impossible. Therefore, we compare our 2D EE strategy against\textit{ the optimal  EE trigger} with layer-wise speed-up calculated by formulas (\ref{eq:threshold}) and (\ref{eq:reduction_layer}). This baseline provides the \textit{supremum speed-up} of all possible layer-wise exit triggers, implying that the additional speed-up achieved with the 2D EE strategy is guaranteed with respect to a range of previous layer-wise EE studies.

The experiments were conducted in two training modes described in Section \ref{sec:training}. For each mode, we compared the results of the optimal layer-wise early exit described above with those of the 2D early exit.
Finally, we also conducted experiments that compare the 2D EE results with the LayerSkip method \cite{Elhoushi_2024} for the Llama-3.1-8B and Llama-3.2-3B models (as the LayerSkip library is not available for Gemma nor Qwen).

\subsection{Datasets}
We used three publicly available datasets for sentiment analysis. 
\begin{enumerate}
    \item Steam gaming Reviews\footnote{\url{https://www.kaggle.com/datasets/andrewmvd/steam-reviews}} (\textit{Steam}), two classes.

    \item The Massive Multilingual Sentiment Corpora\footnote{\url{https://huggingface.co/datasets/Brand24/mms}} (\textit{MMS}) of reviews in multiple languages and topics, three classes.

    \item Amazon Reviews for SA fine-grained\footnote{\url{https://www.kaggle.com/datasets/yacharki/amazon-reviews-for-sentianalysis-finegrained-csv}} (\textit{Amazon-5}), Amazon product reviews, five classes.
\end{enumerate}
From each dataset, we randomly chose a sentiment-balanced English subset of 15K samples (10K train, 5K test, English only) restricted to samples with at least 10 sentences.

\subsection{Language models}

ModernBERT is the encoder-only model released in 2024, and we use it as a baseline for our experiments. The Llama, Gemma, and Qwen families are popular and widely used medium-sized decoder-only LLM. We have chosen four of these models, sized from 3B to 8B parameters. 

\medskip
\begin{small}    \centering
    \begin{tabular}{lll}
    \toprule
      \bf Model name & \bf Layers  & \bf Model card\\
         \midrule
ModernBERT Large & 28 & {https://huggingface.co/answerdotai/ModernBERT-large} \\
Llama-3.1-8B Instruct & 32 & {https://huggingface.co/meta-llama/Llama-3.1-8B-Instruct}\\
Llama-3.2-3B Instruct&  28 & {https://huggingface.co/meta-llama/Llama-3.2-3B-Instruct}\\
Gemma-3n-E4B  &  35 &  {https://huggingface.co/google/gemma-3n-E4B-it}\\
Qwen2.5-7B  &  28 &  https://huggingface.co/Qwen/Qwen2.5-7B-Instruct\\
         \bottomrule
    \end{tabular}
\end{small}

\subsection{Implementation}

The experiments were conducted on a single-node DGX Station A100 with 1× AMD EPYC 7742 (64 cores / 128 threads), 4× Nvidia A100 SXM4 GPUs with 40 GB of HBM2 each (NVLink-connected), and approximately 512 GB of system RAM.
We used the same training setup for all models. The training of one adapter took 2.7 min and, as the adapters are independent, all can be trained in parallel. We used the following setting:
\begin{itemize}
\item Optimiser: Adam with a learning rate $1 \times 10^{-5}$
\item Batch size: 128
\item Epochs: 20
\end{itemize}

For fine-tuning, including ModernBERT and LayerSkip implementations, the LoRA configuration included a rank of 64 and a scaling factor (alpha) of 16, resulting in a training time (Llamma-3.1-8B) of 66.9 min:
\begin{itemize}
\item Optimiser: AdamW with a learning rate $1 \times 10^{-4}$
\item Batch size: 16
\item Epochs: 2
\end{itemize}

For the layer-wise early exit, the model was fed the whole input example, and classifiers at each layer calculated their predictions from the embedding of the last sentence of the example (which aggregated information from the entire example thanks to the attention mechanism).

The 2D early exit inference was implemented by Algorithm \ref{alg:early-exit}, controlled by two hyperparameters $\tau_{ignore}$ and $\tau_{acc}.$ The algorithm is designed for downstream task inference, where the hyperparameters are already pre-set. A systematic exploration of hyperparameter space ($\approx 250$ settings) was needed to explain the behaviour of the algorithm for all 24 combinations of dataset/model/training\_mode -- see Fig. \ref{fig:llama3_hyperparameter_heatmap} for an example. To optimise time and energy, we pre-computed and stored embeddings for each sentence of all testing samples and all layers of each model, along with the corresponding metadata. The embeddings were used to compute the probability vectors $[p_0, p_1, p_2, \ldots]$ in the output of the trained classifiers, which were then processed by Algorithm \ref{alg:early-exit}. This sped-up the experiments by $\approx$ two orders of magnitude.

\section{Results}\label{sec_results_summary}

We compared the efficiency of the 2D early exit strategy with other EE methods tested, based on the metrics detailed in Section \ref{sec_metrics}. Generally, the 2D early exit strategy achieved a substantial speed-up increase across different model architectures for simpler tasks, while its results were inconclusive for the most complex Amazon-5 dataset. The 2D EE advantage depends on the task complexity, the training protocol (adapter-only training versus model-fine-tuning), and the allowed accuracy loss, as detailed in Sec. \ref{sec:results_analysis}. 

\subsection{Baseline results}
\label{sec_baseline}

\begin{figure}[t]
    \includegraphics[width=1\linewidth]{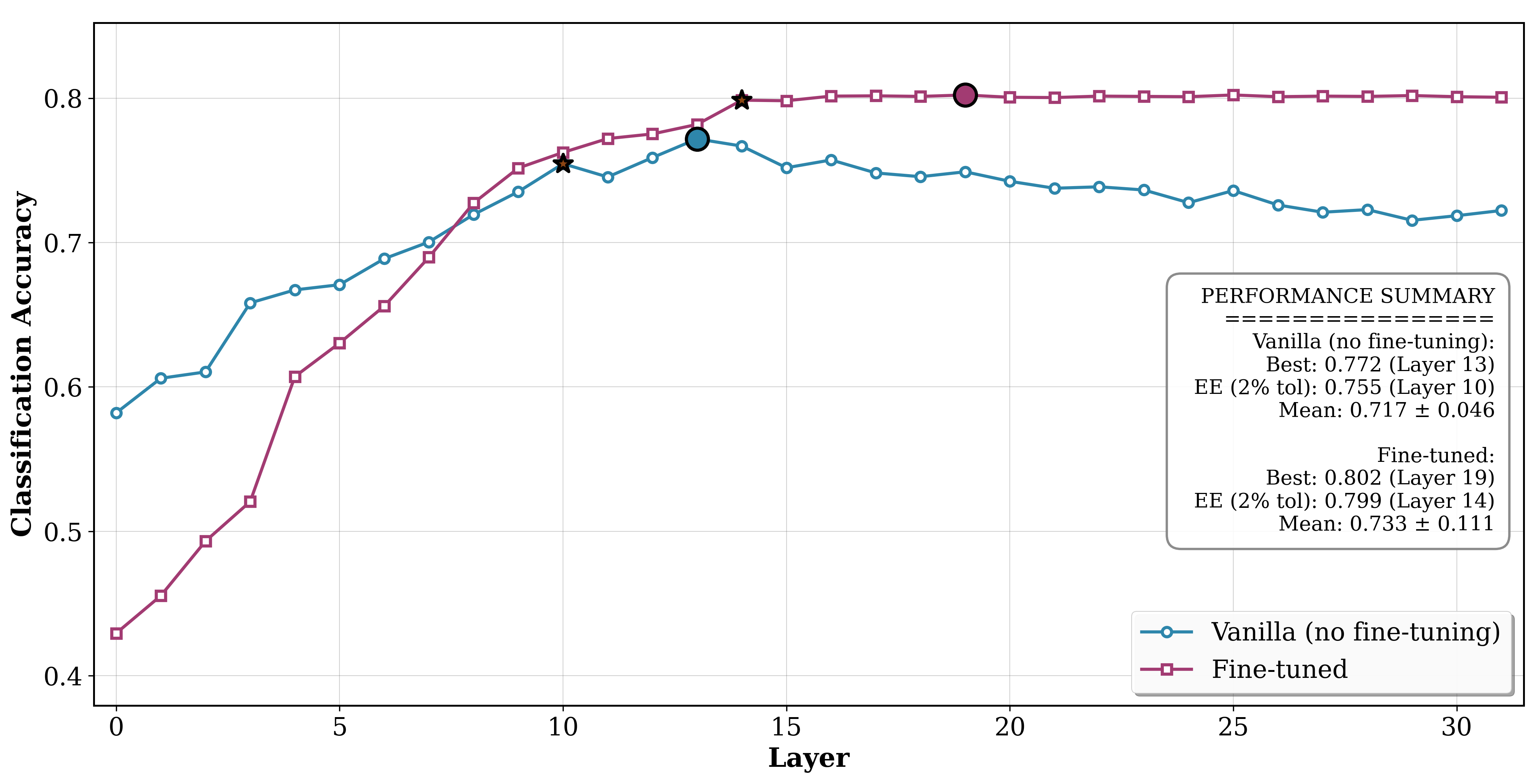}
    \caption{Layer-wise accuracy of LLama-3.1-8B, vanilla model (above) and fine-tuned model (bellow), averaged on the MMS test dataset. Full circles mark the layers with the best accuracy. Stars mark the first layers with the accuracy 2\% below maximum.}
    \label{fig:llama3_MMS_layer-wise}
\end{figure}

To put results into context with numerous studies using BERT-based models, we compared the accuracy of ModernBERT Large with the mid-size LLMs used in our 2D EE study. The results of the fine-tuned models in the three sentiment classification datasets are listed in Table \ref{tab:BERT_accuracy}. No early exit was used.

\begin{table}[t]
    \centering
    \begin{tabular}{|l|c|c|c|c|c|}
    \hline
      \textbf{Model}   & ModernBERT & Llama-3.1 & Llama-3.2 & Gemma-3n & Qwen2.5 \\
      \textbf{Dataset} &  Large     & 8B  &  3B &  E4B &  7B\\
      \hline
      \textbf{Steam}     & 0.87 & 0.97 & 0.96 & 0.96 & 0.96 \\
      \textbf{MMS}       & 0.78 & 0.80 & 0.79 & 0.78 & 0.79 \\
      \textbf{Amazon-5}  & 0.62 & 0.65 & 0.63 & 0.63 & 0.64 \\
      \hline
    \end{tabular}
   \caption{Accuracy of the fine-tuned ModernBERT Large versus other models tested.}
    \label{tab:BERT_accuracy}
\end{table}
%

As an early exit speed-up baseline, we use the optimal idealised layer-wise early exit. Fig. \ref{fig:llama3_MMS_layer-wise} shows the typical layer-wise accuracy curves for vanilla and fine-tuned LLMs (no early exit). In vanilla models, the layer-wise accuracy usually reaches its maximum approximately halfway through the model depth and then slightly decreases. Fine-tuning increases the accuracy in all models as expected, but also flattens the layer-wise accuracy curves in the upper layers. To view this phenomenon from another perspective, Fig. \ref{fig:llama3_embedding_distances} shows the cosine distances between embeddings in consecutive layers of Llama-3.1-8B. 

\subsection{Results of adapter-only training}\label{sec_results_vanilla}

The results are summarised in Table \ref{tab:Summary-vanilla}. In some cases marked with *, the 2D EE strategy did not reach the accuracy threshold (\ref{eq:threshold}) for a loss of 2\%.
For simpler classification tasks (Steam with 2 classes and MMS with 3 classes), the 2D approach consistently outperformed the layer-wise early exit by factors of 1.4-2.3$\times$ (Steam) and 1.1-2.7$\times$ (MMS). 
For the most challenging dataset (Amazon-5 with 5 classes), 2D EE dominated only for Gemma and Qwen models, providing 1.1-1.5 speed-up improvement, while it was unfavourable for Llama models.

\begin{table}[h!]
    \centering
    \begin{tabular}{|l|l|c|rcc|rcc|}\hline
           &  &  &\multicolumn{3}{c|}{\textbf{Accuracy loss 2\%}} & \multicolumn{3}{c|}{\textbf{Accuracy loss 4\%}} \\
          \textbf{Dataset}&\textbf{Model} &  \textbf{Acc} &\multicolumn{2}{c}{\textbf{layer-wise EE}}   & \textbf{2D EE} & \multicolumn{2}{c}{\textbf{layer-wise EE}}&\textbf{2D EE}\\
    &&  &  Layer&  Spd-up& Spd-up
 & Layer& Spd-up&Spd-up
\\\hline
          &Gemma-3n-E4B&  0.94
&   14&  2.3& \textbf{4.2}
 &  11& 2.9
&\textbf{6.4}\\
          &Llama-3.1-8B& 0.95
&  9&  3.2&  \textbf{5.8}
 & 7& 4.0&\textbf{9.2}\\
          Steam&Llama-3.2-3B&  0.94
&  9&  2.8& \textbf{5.3}
 & 7& 3.5
&\textbf{7.5}\\
          &Qwen2.5-7B&  0.95&  12&  2.2& \textbf{3.1}
 & 10& 2.5&\textbf{5.1}\\\hline
 & Gemma-3n-E4B& 0.74
& 14& 2.3&\textbf{4.6} & 12& 2.7
&\textbf{6.2}\\
 & Llama-3.1-8B& 0.77
& 11& 2.7&\textbf{3.3}
 & 9& 3.2
&\textbf{4.3}\\
 MMS& Llama-3.2-3B& 0.76
&  10 & 2.5 & * & 9& 2.8
&\textbf{3.2}\\
 & Qwen2.5-7B& 0.76& 13& 2.0&\textbf{3.7}
 & 13& 2.0&\textbf{5.0}\\\hline
 & Gemma-3n-E4B& 0.55
& 15& 2.2&\textbf{2.5} & 14& 2.3
&\textbf{3.5}\\
 & Llama-3.1-8B& 0.61
& 10& \textbf{2.9}& * & 9& \textbf{3.2}
&1.6
\\
 Amazon-5& Llama-3.2-3B& 0.59
& 10& \textbf{2.5}& * & 9& \textbf{2.8}
&1.6
\\
 & Qwen2.5-7B& 0.56& 14& 1.9&\textbf{2.1}
 & 13& 2.0 &\textbf{2.9}\\\hline
    \end{tabular}
    \caption{Results of the 2D early exit strategy in vanilla models. The cases when 2D EE did not reach the 2\% accuracy loss threshold are marked by *.}
    \label{tab:Summary-vanilla}
\end{table}

\subsection{Results of fine-tuned models}\label{sec_results_FT}

Fine-tuning (Table \ref{tab:Summary-FT}) significantly reduced the advantage of the 2D EE strategy, which remained clearly superior for the simplest Steam dataset by a factor of 1.4-1.7$\times.$. In MMS with 3 classes, the advantage was still 1.05-1.25$\times$ for 4\% accuracy loss, but almost vanished for 2\% loss. Finally, the layer-wise early exit outperformed the 2D EE method on the Amazon-5 dataset for both 2\% and 4\% accuracy loss. This suggests that, while adapter-only training preserves the layer-wise accuracy variations beneficial for 2D early exit, model fine-tuning flattens these variations (Fig. \ref{fig:gemma_heatmap_sentences}), hence reducing the 2D advantage despite improving overall accuracy. Further analysis is provided in the ablation study, Sect. \ref{sec_ablation}.

\begin{table}[h!]
    \centering
    \begin{tabular}{|l|l|c|rcc|rcc|}\hline
           &  &  &\multicolumn{3}{c|}{\textbf{Accuracy loss 2\%}} & \multicolumn{3}{c|}{\textbf{Accuracy loss 4\%}} \\
          \textbf{Dataset}&\textbf{Model} &  \textbf{Acc} &\multicolumn{2}{c}{\textbf{layer-wise EE}}   & \textbf{2D EE} & \multicolumn{2}{c}{\textbf{layer-wise EE}}&\textbf{2D EE}\\
    &&  &  Layer&  Spd-up& Spd-up
 & Layer& Spd-up&Spd-up
\\\hline
          &Gemma-3n-E4B&  0.96
&  13&  2.5& \textbf{3.2}
 &  12& 2.7
&\textbf{4.2}
\\
          &Llama-3.1-8B& 0.97
&  9&  3.2&  \textbf{4.3}
 & 7& 4.0
&\textbf{6.9}
\\
    Steam& Llama-3.2-3B&  0.96
&  8&  3.1& \textbf{4.1}
 & 7& 3.5
&\textbf{5.8}
\\
          &Qwen2.5-7B& 0.96&  10&  2.5& \textbf{3.2}
 & 9& 2.8&\textbf{4.5}\\\hline
 & Gemma-3n-E4B&  0.78
&  19&  1.8& \textbf{1.9}
 & 15& 2.2
&\textbf{2.3}
\\
 & Llama-3.1-8B& 0.80
& 14& 2.1&\textbf{2.2}& 10& 2.9
&\textbf{3.2}
\\
 MMS& Llama-3.2-3B& 0.79
& 12& \textbf{2.2}&2.1
 & 10& 2.5
&\textbf{2.9}
\\
 & Qwen2.5-7B& 0.79& 15& 1.8&\textbf{1.9}
 & 13& 2.0&\textbf{2.5}\\\hline
 & Gemma-3n-E4B& 0.63
& 18& \textbf{1.8}&1.6
 & 14& \textbf{2.3}
&1.9
\\
 & Llama-3.1-8B& 0.65
& 14& \textbf{2.1}&1.6
 & 12& \textbf{2.5}
&2.1
\\
 Amazon-5& Llama-3.2-3B& 0.63
& 12& \textbf{2.2}&1.6
 & 10& \textbf{2.5}
&2.1
\\
 & Qwen2.5-7B& 0.64& 14& \textbf{1.9}&1.4 & 14& \textbf{1.9}&1.8\\\hline
    \end{tabular}
    \caption{Results of the 2D early exit strategy in fine-tuned models.}
    \label{tab:Summary-FT}
\end{table}

\subsection{LayerSkip experiments}\label{sec_layerSkip}

The experiments were carried out for the Llama-3.1-8B and Llama-3.2-3B models, which are supported by the available LayerSkip implementation. Unlike the original paper \cite{Elhoushi_2024}, where the exit layer is manually set as one of the hyperparameters, we let LayerSkip pass through all layers and stored the accuracies per layer.
Its best accuracy among all layers was in line with our layer-wise EE results for all three datasets, confirming the robustness of the methodology used. For comparison with the 2D EE approach (Tab. \ref{tab:layerskip}), we took the first layer in which LayerSkip matched the accuracy threshold (\ref{eq:threshold}) and calculated the corresponding speed-up (\ref{eq:reduction_layer}).  The speed-up of the 2D EE strategy was clearly superior for the Steam and MMS datasets. LayerSkip dominated for the Amazon-5 dataset with an accuracy loss of  2\%, but 2D EE matched or outperformed LayerSkip for 4\% loss.
    
\begin{table}[h!]
    \centering
    \begin{tabular}{|l|l|c|rcc|rcc|}\hline
           &  &  &\multicolumn{3}{c|}{\textbf{Accuracy loss 2\%}} & \multicolumn{3}{c|}{\textbf{Accuracy loss 4\%}} \\
          \textbf{Dataset}&\textbf{Model} &  \textbf{Acc} &\multicolumn{2}{c}{\textbf{LayerSkip}}   & \textbf{2D EE} & \multicolumn{2}{c}{\textbf{LayerSkip}}&\textbf{2D EE}\\
    &&  &  Layer&  Spd-up& Spd-up
 & Layer& Spd-up&Spd-up\\
\hline
Steam &Llama-3.1-8B& 0.97
&  11&  2.9&  \textbf{4.3}
 & 10& 3.2&\textbf{6.9}\\
    & Llama-3.2-3B&  
0.96&  11&  2.5& \textbf{4.1}
 & 10& 2.8&\textbf{5.8}\\\hline
MMS & Llama-3.1-8B& 0.80
& 15& \textbf{2.1}
& \textbf{2.1}
& 15& 2.1&\textbf{3.2}
\\
 & Llama-3.2-3B& 
0.79
& 15& 1.9& \textbf{2.1}
& 15& 1.9&
\textbf{2.9}
\\\hline
Amazon-5 & Llama-3.1-8B& 0.65
& 16& \textbf{2.0}& 1.6& 15& \textbf{2.1}&\textbf{2.1}
\\
 & Llama-3.2-3B& 
0.63& 15& \textbf{1.9}& 
1.6
 & 15& 1.9&
\textbf{2.1}
\\\hline
    \end{tabular}
     
    \caption{Results of LayerSkip compared to the 2D EE technique.}
    \label{tab:layerskip}
\end{table}

\subsection{Analysis of results} 
\label{sec:results_analysis}

The advantage of the 2D EE strategy over layer-wise early exit (or the LayerSkip method) in our experiments depends on several factors. 

\subsubsection*{Task complexity} 
The 2D EE dominated in simpler tasks (Steam and MMS dataset with classification to 2 or 3 classes, respectively), but layer-wise EE proved more efficient for the most challenging task (Amazon-5 with 5 classes), except for two test cases, suggesting that 2D strategies excel when semantic information accumulates predictably across sentences. 

\subsubsection*{Accuracy loss} 
The chosen loss of 2\% balanced performance preservation with speedup across all tested configurations, while the loss of 4\% favoured the 2D EE methods. This is attributed to the fact that 2D EE provides much 
more possible exit points, resulting in a graceful accuracy decay when cutting both the number of processing layers and the processed part of the input. 

\subsubsection*{Fine-tuning} 
Fine-tuning consistently decreased the advantage of 2D EE over layer-wise EE across all three datasets and all models tested. We leave a deeper analysis of this phenomenon to the ablation study in Sect. \ref{sec_ablation}.

\section{Discussion}
\label{sec_discussion}

\subsection{Ablation study}
\label{sec_ablation}

To understand the individual contributions of different components in our 2D early exit framework, we conducted an ablation study on the Gemma-3n-E4B model using the MMS dataset. We compared four configurations: (i) vanilla model with adapter-only training and layer-wise early exit, (ii) same model with 2D early exit, (iii) fine-tuned model with layer-wise early exit, and (iv) same model with 2D early exit.

Table~\ref{tab:ablation} shows the results for accuracy loss of 2\% . The baseline configuration (i) achieves 2.3$\times$ speed-up through layer-wise early exit alone. Adding sentence-wise trimming in configuration (ii) increases the speed-up to 4.6$\times$, demonstrating that the 2D coordination provides a 2$\times$ multiplicative benefit beyond layer-wise exiting. However, model fine-tuning (iii) and (iv), while improving overall accuracy (0.78 vs. 0.74), provides a speed-up of 1.8 / 1.9 for layer-wise / 2D early exit, respectively, reducing the multiplicative benefit of 2D EE to 1.06$\times$.

\begin{table}[h]
\centering
\begin{tabular}{|lcc|}
\hline
Configuration & Max. accuracy& Speed-up \\
\hline
(i) Adapter + Layer-wise EE & 0.74 & 2.3$\times$ \\
(ii) Adapter + 2D EE & 0.74 & 4.6$\times$ \\
(iii) Fine-tuned + Layer-wise EE & 0.78 & 1.8$\times$ \\
(iii) Fine-tuned + 2D EE & 0.78 & 1.9$\times$ \\
\hline
\end{tabular}
\caption{Ablation study results on Gemma-3n-E4B with MMS dataset (accuracy loss of 2\%).}
\label{tab:ablation}
\end{table}

\begin{figure}
    \centering
    \includegraphics[width=\linewidth]{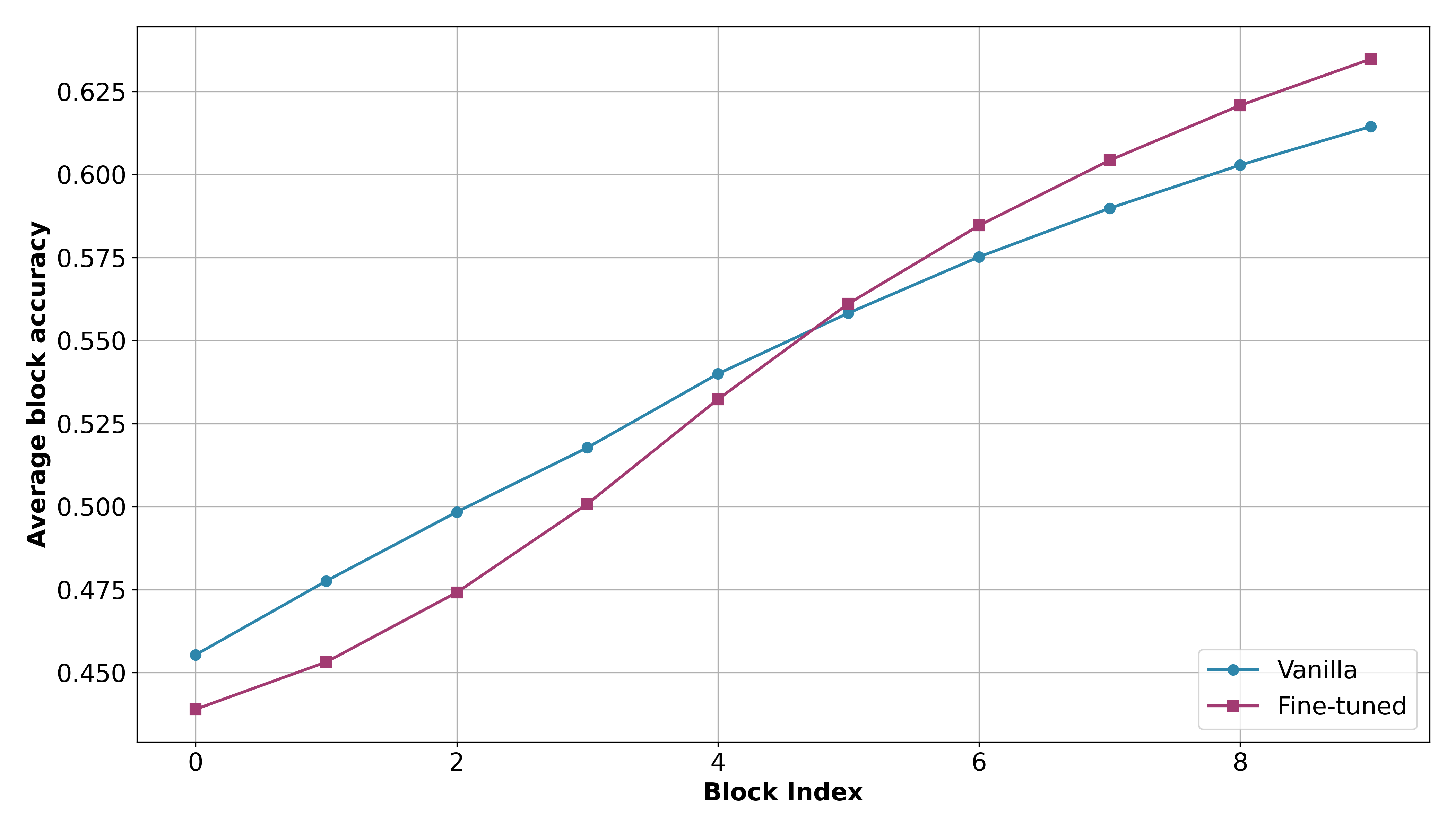}
    \caption{Average block accuracies of 2D early exit for Gemma-3n-E4B on the MMS sub-dataset of samples with 10 sentences.}
    \label{fig:sentence_wise_graph}
\end{figure}

To analyse the role of fine-tuning, we calculated heatmaps of accuracy along the layer and sentence axis for Gemma-3n-E4B, both vanilla and fine-tuned (\ref{fig:gemma_heatmap_sentences}) on an MMS sub-dataset of samples with 10 sentences (7100 samples). Fig. \ref{fig:sentence_wise_graph} compares the average block-wise accuracy curves based on these data for 2D EE progression blocks defined in Sect. \ref{subsec_inference}. Rather surprisingly, while fine-tuning increased accuracy in later layers and sentences, it simultaneously decreased it in earlier layers and sentences, despite the fact that the fine-tuning loss function (\ref{eq_FT_loss}) progressively penalises loss in early layers and sentences. 

\subsection{Hyperparameter setting}

The hyperparameters used during training include the weights $\lambda_i$ of the losses of individual layers in the aggregate loss function (\ref{eq_FT_loss}). While \cite{elbayad2020depth,Elhoushi_2024} use a quadratic increase in weights towards higher layers, we experimentally verified that a simpler setting described in Section \ref{sec:training} provided comparable results, indicating the robustness of the weights. 

Inference requires tuning two hyperparameters: $\tau_{ignore}$ (confidence threshold for accumulation) and $\tau_{acc}$ (acceptance threshold for early stopping). Analysis of the hyperparameter space, illustrated in Figure \ref{fig:llama3_hyperparameter_heatmap}, reveals a low sensitivity to hyperparameters: well-behaved optimisation landscapes with broad and clear global maxima, making efficient search strategies viable. A simple hill-climbing or coarse-to-fine grid search can identify near-optimal configurations within 20-30 evaluations. The optimal values tend to cluster predictably: $\tau_{ignore}$ typically ranges from 0.3 to 0.5 (filtering low-confidence predictions), while $\tau_{acc}$ scales with the complexity of the task. 

\subsection{Comparison with other works} 

Our results align with and extend existing findings. DeeBERT achieved $\approx 40$\% inference reduction in the GLUE SST-2 task with $<4\%$ accuracy loss, in particular 44-47\% reduction in SST-2 with $\approx 2\%$ accuracy loss, comparable to our experiment with Steam dataset (3.2-4.3$\times$ speed-up, i.e., 69-77\% reduction, $<$2\% loss). The comparison with LayerSkip showed an advantage of 2D EE $\approx 1.6\times$ for the Steam dataset (2\% accuracy loss) and $\approx 1.8\times$ for MMS (4\% loss). Unlike methods requiring architectural modifications (LayerSkip, MatFormer) or reinforcement learning (ConsistEE, GREEN-CODE), our approach remains model-agnostic and training-efficient, requiring only lightweight classification adapters. The key distinction is our simultaneous optimisation across both layer and token dimensions, which -— as demonstrated -— can provide multiplicative rather than additive benefits when task complexity permits early semantic convergence. 

\subsection{Limitations}

\subsubsection*{Datasets}
We used three public datasets of customer reviews with different difficulty (2 / 3 / 5 classes). Consequently, the fine-tuned models returned average test set accuracies of 0.96 / 0.79 / 0.64, with only slight variations between models. This choice of datasets allowed us to demonstrate how 2D EE efficiency depends on the complexity of the task. We could not use some popular benchmarks such as GLUE, since longer inputs ($\geq$10 sentences) are needed to demonstrate the 2D EE strength. Experiments with datasets beyond sentiment classification will be needed to assess general applicability. 

\subsubsection*{Baselines}
To avoid comparison with numerous previous works using various training and inference mechanisms, which necessarily leads to methodological inconsistency and incompleteness, we generalised layer-wise EE studies to an \textit{optimal early exit} using an exit layer with the best possible speed-up (Section \ref{sec_experiments}). This allowed for a principal comparison of the 1D (layer-wise) and 2D (layer-wise and token-wise) early exit philosophy. In addition, we included a direct comparison of the results with an influential recent study \cite{Elhoushi_2024}.

\subsubsection*{Metrics}
Early exit trades a substantial speed-up (and energy reduction) for a small output quality drop. The choice of accuracy metrics suffices for this purpose, as in other similar studies. The inference speed-up was expressed as the ratio of abstract operations (processed layers/input sentences) used in the full and the exiting model, which is machine-dependent model-agnostic. We do not include wall-clock timing results because a fair runtime evaluation would require an optimised custom CUDA kernel for the proposed early-exit computation. A naive implementation would not faithfully represent the method’s true efficiency, as the measured latency would be dominated by implementation overhead rather than by the underlying reduction in computation. Developing such a kernel is outside the scope of this paper, so we restrict our evaluation to theoretical/computational analysis.

\subsection{Orthogonality with other methods}

The 2D EE strategy can be combined with a range of complementary methods to further enhance efficiency. One option is to integrate 2D EE with \textit{Mixture of Experts} (MoE) architectures \cite{shazeer2017outrageously}, which dynamically route input through a sparse set of specialised sub-networks. 
\textit{Matryoshka} representation learning \cite{NEURIPS2022_matryoshka} organises representations into nested levels of complexity. Among the models we tested, Gemma-3n-E4B uses the Matryoshka architecture, and therefore our results confirm its orthogonality with the 2D EE technique.
Similarly, \textit{quantization methods} \cite{dettmers2023qlora} can allow intermediate exits to operate with low precision when used alongside 2D EE.
\textit{Pruning techniques} \cite{ma2023llm} can be applied to remove redundant weights or neurons. Combined with 2D EE, pruning can further lower computational requirements by simplifying both the early and late stages of the model. 
Finally,\textit{ knowledge distillation} \cite{jiao2020tinybert} offers a compelling way to train intermediate exit layers. By distilling knowledge from the full model into these layers, one can boost the accuracy of early predictions in both layer-wise and token-wise levels.


\section{Conclusion and future work}
\label{sec_conclusion}

We have introduced a novel two-dimensional early exit strategy that simultaneously coordinates layer-wise and sentence-wise exiting for classification tasks in large language models. By processing input incrementally sentence-by-sentence while progressively activating deeper layers, our method achieves computational savings that exceed those possible through layer-wise or token-wise optimisation alone. 

Our experimental evaluation across four state-of-the-art LLMs (Gemma-3n 4B, Llama-3.1-8B, Llama-3.2-3B, Qwen-2.5 7B) on three sentiment classification datasets demonstrates substantial additional speed-ups against optimal layer-wise early exit with vanilla models: up to 2.3× for both binary classification and a three-class problem, and up to 1.5× for complex multi-class problems (Amazon-5 with 5 classes). The 2D advantage persists, though diminished, also with fine-tuned models (up to 1.7× for two classes and up to 1.3× for three classes). However, the 2D advantage disappears on Llama models for the Amazon-5 task, indicating that 2D strategies excel primarily when semantic information accumulates predictably across sentences.

An intriguing finding is that fine-tuning, while improving overall accuracy, paradoxically reduces the 2D early exit advantage by flattening accuracy variations across early layers and partial inputs. This suggests an unexplored tension between optimising for final accuracy and preserving the gradient structures that enable effective early exiting. The heatmaps in Fig. \ref{fig:gemma_heatmap_sentences} and the graph in Fig. \ref{fig:sentence_wise_graph} illustrate this phenomenon, revealing that fine-tuning increases accuracy in later processing stages while simultaneously decreasing it in earlier ones—despite the loss function explicitly encouraging early predictions.

Future work should address several directions. First, developing fine-tuning strategies that explicitly preserve or enhance accuracy gradients across both dimensions could unlock the full potential of 2D early exit. Techniques for stabilising layer-wise semantics, such as GER-steer \cite{jiang2026globalevolutionarysteeringrefining}, might be helpful. Second, extending the approach beyond sentiment classification to tasks such as topic categorisation, intent detection, or question answering would test its generalisability. Third, integrating reinforcement learning to enable adaptive, per-sample layer skipping patterns could provide even more flexible resource allocation. Finally, combining 2D early exit with orthogonal techniques such as quantisation, pruning, or mixture-of-experts architectures could yield multiplicative efficiency gains across multiple optimisation dimensions.

\section*{Acknowledgements}
The authors thank Lou Nováků for carrying out experiments with ModernBERT.

\section*{Declarations}
\subsection*{Funding}
This article was produced with the financial support of the European Union under the REFRESH – Research Excellence For REgion Sustainability and High-tech Industries project number CZ.10.03.01/00/22\_003/0000048 via the Operational Programme Just Transition, and under the: Biography of Fake News with a Touch of AI: Dangerous Phenomenon through the Prism of Modern Human Sciences project no.: CZ.02.01.01/00/23\_025/0008724 via the Operational Programme Jan Ámos Komenský. It was also supported by the Silesian University in Opava under the Student Funding Plan, project SGS/9/2024.

\subsection*{Conflict of interest}
The authors declare that they have no conflict of interest.

\subsection*{Data availability}
Project (continuously updated): \url{https://github.com/irafm-llm/2D_early_exit_inference} \\
Datasets: \url{https://huggingface.co/datasets/irafm-llm/2D_early_exit_inference}


\bibliographystyle{plainnat}
\bibliography{references}

\newpage
\begin{appendices}

\section{Splitting the text into sentences}
\label{app_sentence_splitting}

The datasets contain one user contribution per sample, such as a product review, which must be split into individual sentences. Due to the informal style of user reviews, simple splitting is insufficient, and regular expressions are required.
Our solution employs two regular expressions. The first is a negative look-behind, preventing splits at positions that are not genuine sentence boundaries. Specifically, it excludes periods in abbreviations (e.g. "Dr." or "etc.") and multiple consecutive periods ("..."). The second regular expression actively identifies true delimiters -- a period followed by whitespace or an exclamation mark or question mark (potentially repeated) followed by whitespace.
In practice, consider the review "Excellent product! It is great. I recommend Dr. Smith." The algorithm correctly identifies sentence boundaries after "product!" and "great." but preserves "I recommend Dr. Smith." as a unit since the abbreviation "Dr." satisfies the negative look-behind conditions. 

\section{Additional explanatory results}

\begin{figure}[h!]
    \centering
    \includegraphics[width=\linewidth,trim=0 0 0 0.9cm,clip]{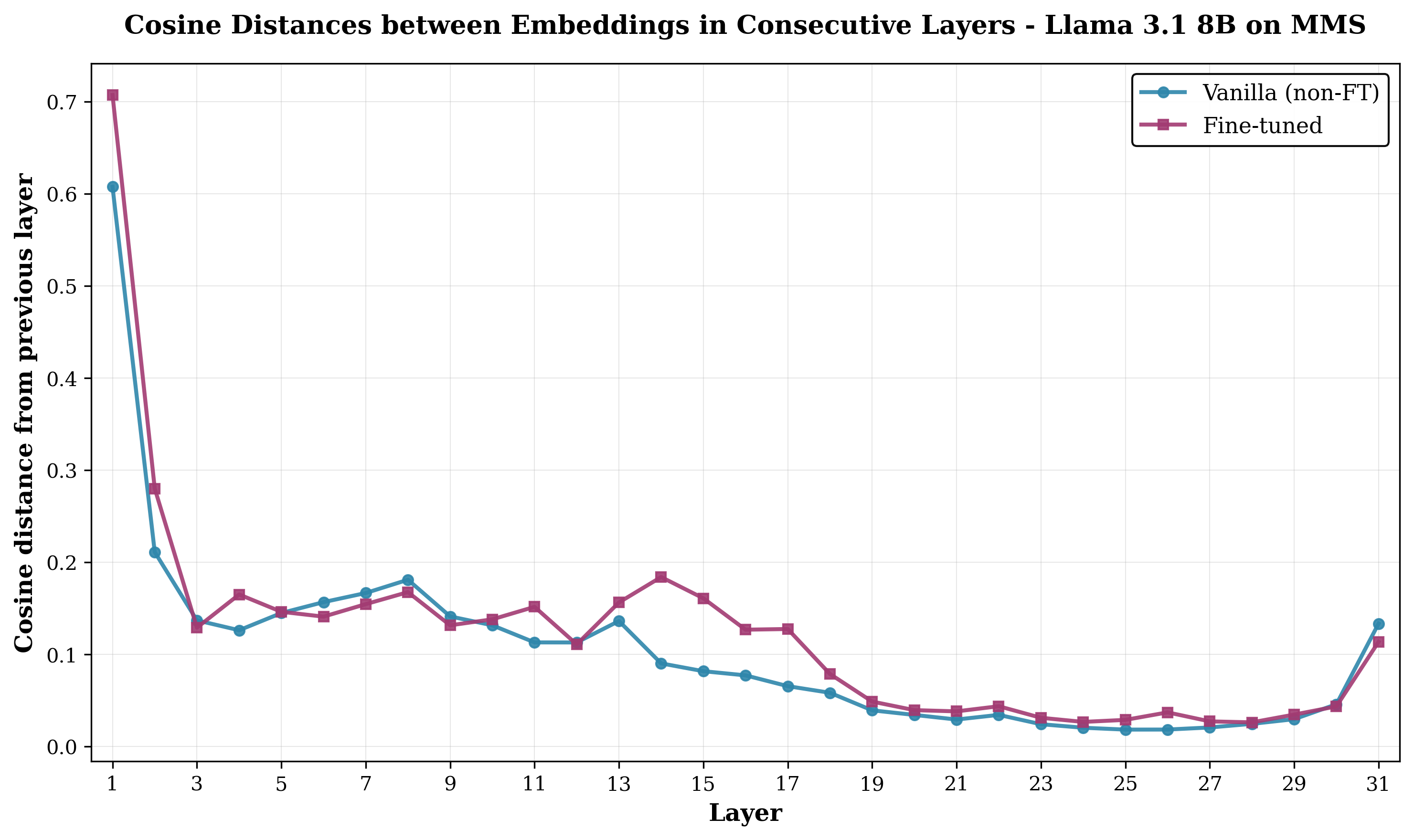}
    \caption{Cosine distances of embedding vectors between consecutive layers of Llama-3.1-8B, averaged over the MMS testing split.}
    \label{fig:llama3_embedding_distances}
\end{figure}

Figure \ref{fig:llama3_embedding_distances} shows cosine distances between embeddings of consecutive layers in Llama-3.1-8B model, in both vanilla and fine-tuned version, using the MMS dataset testing split. The local peaks visible between layers 13-15 correspond to the optimal exit points with the 2\% accuracy loss in Tables \ref{tab:Summary-vanilla} and \ref{tab:Summary-FT}.

\begin{figure}
    \centering
    \vspace*{-2cm}
    \includegraphics[width=.85\linewidth,trim=0 0 0 0.8cm,clip]{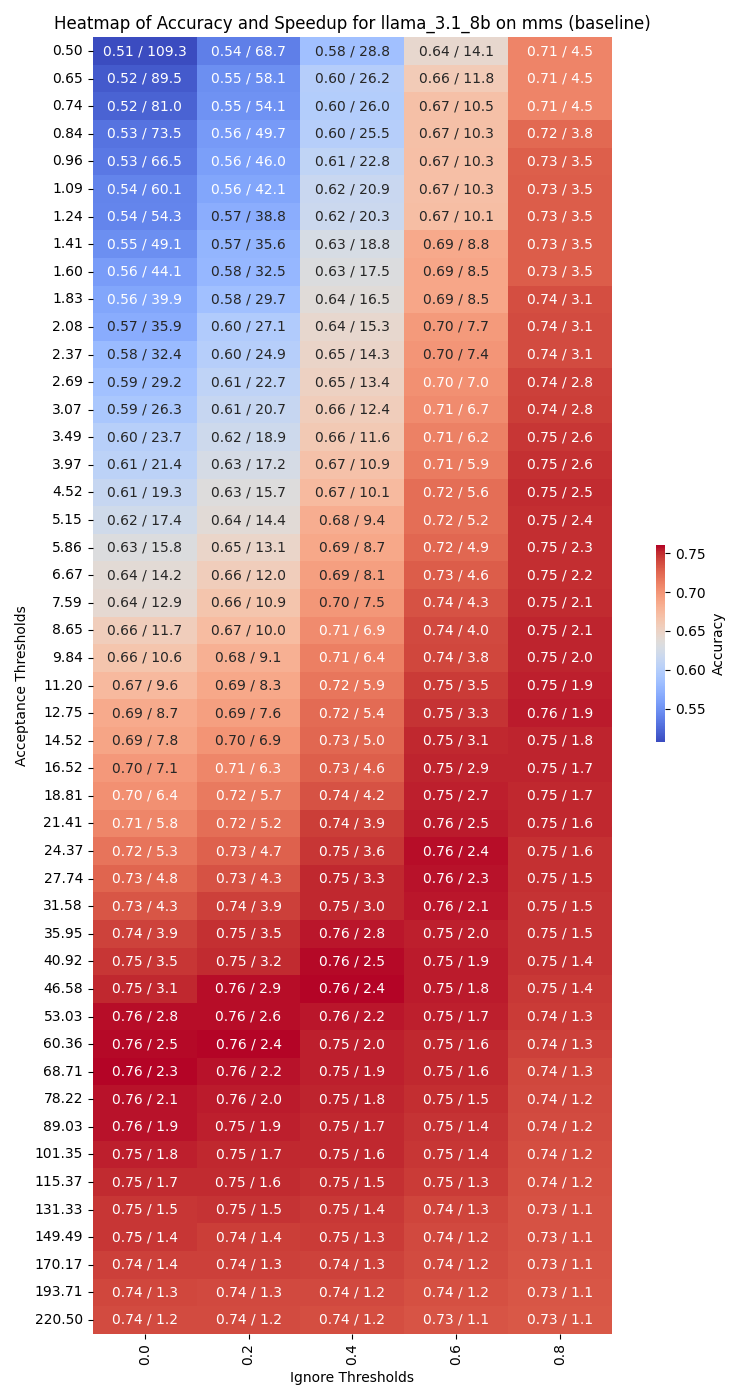}
    \caption{Accuracy and speedup heatmap of the 2D early exit strategy for Llama-3.1-8B on the MMS dataset, no fine-tuning.}
    \label{fig:llama3_hyperparameter_heatmap}
\end{figure}

Figure \ref{fig:llama3_hyperparameter_heatmap} displays the accuracy/speed-up heatmap of the 2D early exit in Llama-3.1-8B model in both vanilla version, using the MMS dataset testing split. The axes are the hyperparameters $\tau_{acc}$ (accuracy threshold) and $\tau_{ignore}$ (ignore threshold). The heatmap show a well-behaved and broad global accuracy maximum, suggesting fast convergence of various hyperparameter search methods. The best speed-up for the 2\% accuracy loss is found for $\tau_{acc} = 27.74,$ $\tau_{ignore} = 0.4.$

\end{appendices}

\end{document}

%% file: Schema_2D_EE.tex
\begin{tikzpicture}[
  scale=0.85,
  transform shape,
  block/.style={
    draw=black!50, thin, rounded corners=1.5pt,
    minimum width=1.15cm, minimum height=0.6cm,
    inner sep=0pt, font=\small\bfseries, text=red!65!black,
    fill=white
  },
  phase/.style={draw=red!65!black, rounded corners=2pt},
  flow/.style={-{Stealth[length=3pt,width=2.5pt]}, thin, black!50},
  jump/.style={-{Stealth[length=3.5pt,width=2.5pt]}, semithick, black!55},
  lbl/.style={font=\small},
]

\def\sx{2.8}   
\def\ly{0.95}  
\def\cg{0.35}  

\pgfmathsetmacro{\ya}{0}
\pgfmathsetmacro{\yb}{\ly}
\pgfmathsetmacro{\yc}{2*\ly + \cg}
\pgfmathsetmacro{\yd}{3*\ly + \cg}
\pgfmathsetmacro{\ye}{4*\ly + 2*\cg}
\pgfmathsetmacro{\yf}{5*\ly + 2*\cg}
\pgfmathsetmacro{\yg}{6*\ly + 3*\cg}
\pgfmathsetmacro{\yh}{7*\ly + 3*\cg}

\pgfmathsetmacro{\routeA}{0.5*(\yd + \ye)}   
\pgfmathsetmacro{\routeB}{0.5*(\yf + \yg)}   

\def\jog{0.55}


\node[block](s0l0) at (0,\ya){0};      \node[block](s0l1) at (0,\yb){1};
\node[block](s0l2) at (0,\yc){2};      \node[block](s0l3) at (0,\yd){3};
\node[block](s0l4) at (0,\ye){8};      \node[block](s0l5) at (0,\yf){9};
\node[block](s0l6) at (0,\yg){18};     \node[block](s0l7) at (0,\yh){19};

\node[block](s1l0) at (\sx,\ya){4};    \node[block](s1l1) at (\sx,\yb){5};
\node[block](s1l2) at (\sx,\yc){6};    \node[block](s1l3) at (\sx,\yd){7};
\node[block](s1l4) at (\sx,\ye){10};   \node[block](s1l5) at (\sx,\yf){11};
\node[block](s1l6) at (\sx,\yg){20};   \node[block](s1l7) at (\sx,\yh){21};

\node[block](s2l0) at (2*\sx,\ya){12}; \node[block](s2l1) at (2*\sx,\yb){13};
\node[block](s2l2) at (2*\sx,\yc){14}; \node[block](s2l3) at (2*\sx,\yd){15};
\node[block](s2l4) at (2*\sx,\ye){16}; \node[block](s2l5) at (2*\sx,\yf){17};
\node[block](s2l6) at (2*\sx,\yg){22}; \node[block](s2l7) at (2*\sx,\yh){23};

\node[block](s3l0) at (3*\sx,\ya){24}; \node[block](s3l1) at (3*\sx,\yb){25};
\node[block](s3l2) at (3*\sx,\yc){26}; \node[block](s3l3) at (3*\sx,\yd){27};
\node[block](s3l4) at (3*\sx,\ye){28}; \node[block](s3l5) at (3*\sx,\yf){29};
\node[block](s3l6) at (3*\sx,\yg){30}; \node[block](s3l7) at (3*\sx,\yh){31};

\node[phase, semithick, inner sep=4pt]  [fit=(s0l0)(s0l1)] {};
\node[phase, semithick, inner xsep=8pt, inner ysep=7pt, yshift=-1pt]  [fit=(s0l0)(s1l3)] {};
\node[phase, semithick, inner xsep=10pt, xshift=-2pt, inner ysep=9pt, yshift=-3pt] [fit=(s0l0)(s2l5)] {};
\node[phase, semithick, inner sep=12pt, xshift=-4pt, yshift=-4pt] [fit=(s0l0)(s3l7)] {};


\foreach \s in {0,1}{
  \draw[flow] (s\s l0)--(s\s l1);
  \draw[flow] (s\s l1)--(s\s l2);
  \draw[flow] (s\s l2)--(s\s l3);
  \draw[flow] (s\s l4)--(s\s l5);
  \draw[flow] (s\s l6)--(s\s l7);
}

\draw[flow] (s2l0)--(s2l1);
\draw[flow] (s2l1)--(s2l2);
\draw[flow] (s2l2)--(s2l3);
\draw[flow] (s2l3)--(s2l4);
\draw[flow] (s2l4)--(s2l5);
\draw[flow] (s2l6)--(s2l7);

\draw[flow] (s3l0)--(s3l1);
\draw[flow] (s3l1)--(s3l2);
\draw[flow] (s3l2)--(s3l3);
\draw[flow] (s3l3)--(s3l4);
\draw[flow] (s3l4)--(s3l5);
\draw[flow] (s3l5)--(s3l6);
\draw[flow] (s3l6)--(s3l7);


\draw[jump] (s0l3.east) -- ++(\jog, 0) |- (s1l0.west);

\draw[jump] (s1l3.north) -- (\sx, \routeA) -- (0, \routeA) -- (s0l4.south);

\draw[jump] (s0l5.east) -- ++(\jog, 0) |- (s1l4.west);

\draw[jump] (s1l5.east) -- ++(\jog, 0) |- (s2l0.west);

\draw[jump] (s2l5.north) -- (2*\sx, \routeB) -- (0, \routeB) -- (s0l6.south);

\draw[jump] (s0l7.east) -- ++(\jog, 0) |- (s1l6.west);

\draw[jump] (s1l7.east) -- ++(\jog, 0) |- (s2l6.west);

\draw[jump] (s2l7.east) -- ++(\jog, 0) |- (s3l0.west);

\pgfmathsetmacro{\lblX}{-1.3}
\node[lbl, anchor=east] at (\lblX, \ya) {Layer~0};
\node[lbl, anchor=east] at (\lblX, \yb) {Layer~1};
\node[lbl, anchor=east] at (\lblX, \yc) {Layer~2};
\node[lbl, anchor=east] at (\lblX, \yd) {Layer~3};
\node[lbl, anchor=east] at (\lblX, \ye) {Layer~4};
\node[lbl, anchor=east] at (\lblX, \yf) {Layer~5};
\node[lbl, anchor=east] at (\lblX, \yg) {Layer~6};
\node[lbl, anchor=east] at (\lblX, \yh) {Layer~7};

\node[lbl, anchor=north] at (0,     -0.9) {sentence~0};
\node[lbl, anchor=north] at (\sx,   -0.9) {sentence~1};
\node[lbl, anchor=north] at (2*\sx, -0.9) {sentence~2};
\node[lbl, anchor=north] at (3*\sx, -0.9) {sentence~3};

\pgfmathsetmacro{\dotgap}{0.55}
\pgfmathsetmacro{\drx}{3*\sx + 0.575 + \dotgap}   
\pgfmathsetmacro{\duy}{\yh + 0.3 + \dotgap}        

\foreach \yy in {\ya, \yb, \yc, \yd, \ye, \yf, \yg, \yh}{
  \node at (\drx+0.1, \yy) {$\cdots$};
}
\node at (0,     \duy+0.2) {$\vdots$};
\node at (\sx,   \duy+0.2) {$\vdots$};
\node at (2*\sx, \duy+0.2) {$\vdots$};
\node at (3*\sx, \duy+0.2) {$\vdots$};
\node at (\drx,      \duy)      {$\cdot$};
\node at (\drx+0.12, \duy+0.12) {$\cdot$};
\node at (\drx-0.12, \duy-0.12) {$\cdot$};

\end{tikzpicture}